\documentclass[journal]{IEEEtran}
\usepackage{cite}
\usepackage{mathtools,amssymb}
\usepackage{siunitx}            % <---
\usepackage{booktabs, makecell} 
\usepackage{amsmath,amssymb,amsfonts}
\usepackage{algorithmic}
\usepackage{hyperref}
\usepackage{graphicx}
\usepackage{float}
\usepackage{subfloat}
\usepackage[font={sf,scriptsize}, % new 
labelfont={bf,color=accessblue},% new 
caption=false]   % new
{subfig} 
\newcommand{\R}{\mathbb{R}}
\usepackage{kantlipsum} % dummy text
\usepackage{dblfloatfix}
\usepackage{subfig}
\usepackage{graphicx}

\begin{document}
\title{Fusion of Satellite Images and Weather Data with Transformer Networks for Downy Mildew Disease Detection}

\author{
	\IEEEauthorblockN{William Maillet}\IEEEauthorrefmark{1},
	\IEEEauthorblockN{Maryam Ouhami}\IEEEauthorrefmark{1},
	\IEEEauthorblockN{Adel Hafiane}\IEEEauthorrefmark{1}
	\IEEEauthorblockA{\\\IEEEauthorrefmark{1}INSA CVL, University of Orléans, PRISME Laboratory EA 4229, Bourges, France} }

\maketitle
%\IEEEtitleabstractindextext{%

\begin{abstract}
Crop diseases significantly affect the quantity and quality of agricultural production. In a context where the goal of precision agriculture is to minimize or even avoid the use of pesticides, weather and remote sensing data with deep learning can play a pivotal role in detecting crop diseases, allowing localized treatment of crops. However, combining heterogeneous data such as weather and images remains a hot topic and challenging task. Recent developments in transformer architectures have shown the possibility of fusion of data from different domains, for instance text-image. The current trend is to custom only one transformer to create a multimodal fusion model. Conversely, we propose a new approach to realize data fusion using three transformers. In this paper, we first solved the missing satellite images problem, by interpolating them with a ConvLSTM model. Then, proposed a multimodal fusion architecture that jointly learns to process visual and weather information. The architecture is built from three main components, a Vision Transformer and two transformer-encoders, allowing to fuse both image and weather modalities. The results of the proposed method are promising achieving 97\% overall accuracy.
\end{abstract}

%}

%\IEEEdisplaynontitleabstractindextext

% \fi
%\IEEEpeerreviewmaketitle

\section{Introduction}
\label{sec:introduction}
Agricultural chemicals such as fungicides and pesticides are increasingly used to avoid and minimize disease damage. Nonetheless, this chemicals overuse has become problematic from an ecological and ethical point of view, while the European Union even aims to halve the use of chemicals by 2030 \cite{Europe2030}. The challenge for farmers in the coming decades is, therefore, to find a way to monitor their lands better in order to maximize localized treatments for the infected crops instead of spraying chemicals on a large scale. Remote sensing is one of the popular tools in precision agriculture as it helps monitoring crops problems such as diseases, weed infestation, lack of water, etc \cite{Weiss2019}.

Satellite imagery and deep learning have been widely used in crop monitoring as they provides applications in multiple areas: lands classification \cite{Persson2018}, yields predictions \cite{Hunt2019}, wildfires management \cite{Castillo2020},  disease detection \cite{santoso2011mapping} and various detection tasks \cite{Themistocleous2020}. In the recent years, multiple technologies have been applied to accomplish these tasks from CNNs \cite{Rosentreter2020} to RNNs \cite{Papadomanolaki2019} and recently transformers \cite{Yuan2022} which are becoming very popular state-of-the-art models, particularly for multimodal data such as images, text,  etc.  

Along with the satellite imagery, the sensor-based weather monitoring devices become an essential part of precision agriculture. Nowadays most of the farmers have weather conditions monitoring facilities. Furthermore, combining weather data with the satellite images is becoming more and more attractive to prevent crop diseases. In this paper, we propose a fusion architecture combining the weather measurements with the satellites images in order to predict downy mildew in the crops. This architecture uses the Vision Transformer (ViT) \cite{ViT} as image features extractor, the fusion with weather data is performed in bottleneck mode. The architecture was trained on images and weather data over a period of two years, and tested on the same type of data over two years, in order to evaluate the performance of the model in predicting the presence of downy mildew. The contributions of this paper are mainly:
\begin{itemize}
	\item A new deep learning-based approach for temporal satellite image generation to handle missing information;
	\item A new multimodal ViT fusion architecture using three-encoder components to integrate heterogeneous data;
	\item An application of disease detection and identification on vine crops using weather and satellite images.
\end{itemize}
The paper is organized as follows. Section 2 provides a review of related works, section 3 describes the proposed approach. Section 4 presents the experiments conducted in this study and the results. The discussion and conclusion are presented in Section 5 and 6 respectively.

%%%%%%%%%%%%%%%%%%%%%%%%%%%%%%%%%%%%%%%%%%%%%%%%%%%%%%%%%%%%%%%%%%%%%%%%%%%%%%%%%%%%%%%%%%

\section{Related Work}
\label{related}
Promising approaches for detecting diseases were proposed in recent years using machine learning and deep learning on weather data \cite{khattab_iot-based_2019, trilles_development_2020}. In addition, Convolution Neural Networks (CNNs) have shown interesting precision results on remote sensing images for disease detection \cite{kerkech_vddnet_2020, 2020soybean, abdulridha2020detecting, poblete2020detection}. However, CNN based models requires a large dataset and the training process takes a lot of time due to the models complexity. Recently, vision transformers were introduced as a better performing solution for computer vision problems.

\subsection{Vision transformers}

Transformers were first proposed by \cite{Vaswani2017}, and they soon revolutionized the Natural Language Processing (NLP) domain. They replaced recurrent neural networks thanks to their shorter training time and parallelism. Yet, multiple architectures using the attention mechanism of the NLP transformers were created shortly after. ViT \cite{ViT} on the other hand, is the most popular transformer used in image classification. It is known for its simple architecture allowing to process images with very little modifications and exploit the full power of transformers. Transformers are becoming popular because they have shown to be more efficient than traditional CNNs. In fact, CNNs can recognize relationship between distant objects because of the convolutional operations whereas the attention mechanisms of transformers allow to get a better global understanding of the images and its intrinsic relationships.

Transformers  have been used for multiple computer vision tasks, namely scene classification \cite{Bazi2021,Zhang2021}, change detection \cite{HaoChen2022}, image segmentation \cite{Xiangkai2021,Xu2021}. ViTs were also used for various tasks in satellites imagery, such as change detection \cite{Horvath2021} and deforestation monitoring  \cite{Kaselimi2022}. The ViTs results were convincing, they even outperformed the classical convolutional architectures.

%-%-%-%-%-%-%-%-%-%-%-%-%-%-%-%-%-%-%-%-%-%

\subsection{Multimodal fusion}
Since the agricultural field provides a rich variety of data from multiple sources, it seems more judicious to merge the variety of data in one model to achieve a better performance. In disease detection, multimodal fusion is still an ongoing area of study \cite{ghamisi2019multisource, ouhami_2021}. Multimodal fusion based on machine learning is able to jointly learn to process different modalities. Depending on the level of data abstraction, different fusion architectures are possible, such as input data level fusion, feature fusion allowing data integration using feature vectors, etc. We can divide deep learning approaches for multimodal fusion into two categories, those based on CNNs and transformer-based architectures.

\subsubsection{CNN based multimodal fusion}
Using multiple modalities to learn is one of the most important topics in machine learning as it is the closest to human learning.  Various multimodal fusion techniques presented in the literature were based on CNNs to extract feature maps from the input images. Those techniques have different applications, namely
in the medical field \cite{liu2020automatic}, in mechanics \cite{jing2017adaptive} and in agriculture specifically data fusion for yield prediction \cite{chu2020end}, land monitoring \cite{song2020identifying}, crop identification \cite{pelletier2019temporal} and disease detection \cite{selvaraj2020detection}.
%----------------------------------

\subsubsection{Multimodal transformers architectures}
Multimodal learning with transformers has been tested in multiple areas, especially in the audiovisual field to join video, language and audio features \cite{Shvetsova2021, NEURIPS2021, Li2020, Botach2022} but also in deepfake detection \cite{Wang2021}, medical imagery synthesis \cite{Dalmaz2022}, etc. Most of these proposed models extract embeddings from the modalities without transformers to make the fusion in a single custom transformer. In fact, they rarely use multiple transformers to extract features. \cite{Nagrani2021} proposed a multimodal architecture using a custom bottleneck transformer to combine features. Nonetheless, this bottleneck architecture is within the transformer and not as a single transformer.
In the satellite imagery areas, \cite{Chen2022} proposed a multimodal fusion architecture using multiple image sources. The modalities features are extracted using LSTM cells and a modified ViT transformer. \cite{Aldahoul123} proposed a multimodal architecture for object detection, using ViT as a feature extractor. Recently, \cite{Roy2022} also proposed modifications of the ViT transformer allowing to use LIDAR data along with multi-spectral images. So far, very little attention has been paid to fusion of satellite images with weather data for crop disease detection.

%%%%%%%%%%%%%%%%%%%%%%%%%%%%%%%%%%%%%%%%%%%%%%%%%%%%%%%%%%%%%%%%%%%%%%%%%%%%%%%%%%%%%%%%%%

\section{Methodology}
\label{method}

%-------------------------------- Architecture Overview Figure -------------------------------%
\begin{figure}[h!]                                                                            
	\centering                                                                                 
	\includegraphics[width=0.6\linewidth]{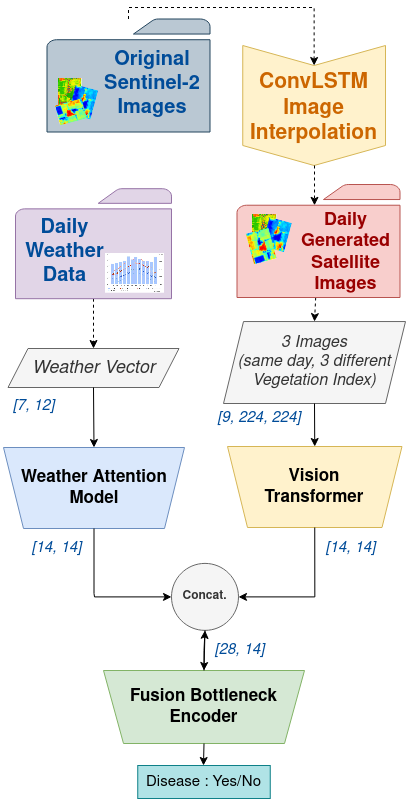}
	\caption{An overview of the proposed approach, from dataset processing to predictions.}
	\label{fig:process}
\end{figure}
%-------------------------------- Architecture Overview Figure -------------------------------%

%proposition de name de l'architecture : ViT multi-Fuse
In this section we describe the proposed method that combines Sentinel-2 satellite images as 2D information, with ground weather data as 1D information. First, we detail the process of increasing the temporal frequency of satellite images by interpolation and deep learning procedure. Second, we describe the fusion architecture between the two heterogenous data. An overview of the entire process flowchart is depicted in   \autoref{fig:process}.

\subsection{Satellite images generation}
\label{sec:Imgen}

\begin{figure*}[ht]
	\centering
	\includegraphics[width=1.\linewidth]{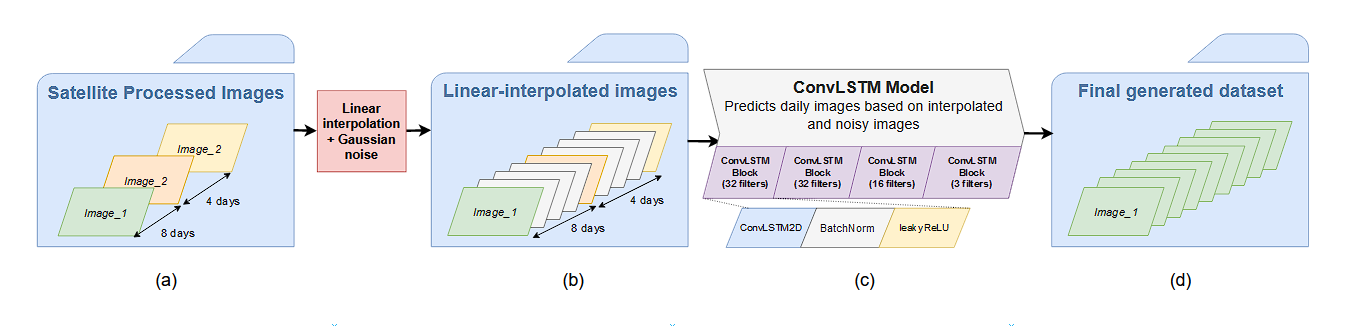}
	\caption{The interpolation process: (a) The satellite image set with temporal gaps (b) being interpolated and noised with Gaussian noise to create an artificial dataset, which is then used (c) to train a ConvLSTM model made of multiple ConvLSTM layers, and (d) the model eventually predicts daily images.}
	\label{fig:convlstm}
\end{figure*}

%The proposed model was tested using Sentinel-2 images. \textcolor{blue}{The main issue in those images is the long revisit cycle of the twin satellites of 5 days. Considering a cloudy region the satellite can only provide one clear image in a period can reach to one month.}
%\textcolor{red}{As the weather data is daily, we would like to have an image per day, in order to match the weather and get daily predictions.}
The Sentinel-2 satellite provides multispectral images of a given location, usually on a 5-day cycle. In cloudy conditions, the temporal resolution between exploitable images can be much lower. As the weather data is daily, it is essential to have one image per day to enable a consistent fusion of 2D/1D information. To overcome this limitation, we propose a data generation method which combines linear interpolation with noise injection and Convolutional Long Short-Term Memory (ConvLSTM). The goal is to train a ConvLSTM model to output the missing intermediate images between two dates. We believe that the ConvLSTM ability to capture temporal and spatial features would give more interesting predictions, compared to only interpolation based on an analytical model. The proposed method is split into two parts: first, we generate linearly interpolated images with gaussian noise, and then we use those images to train a ConvLSTM model outputting the final images that we use for our fusion model (see \autoref{fig:convlstm}).

%--------------------------------------- ConvLSTM Figure --------------------------------------%
%--------------------------------------- ConvLSTM Figure --------------------------------------%
\subsubsection{Temporal interpolation of images}
To generate the training images, we use linear interpolation and gaussian noise. Let $I_{n}$ and $I_{n+k}$ be two real images  (matrices) captured $k$ days apart. To generate all the missing images from day $n+1$ to day $n+k-1$, we use the  \autoref{eq:interpolation}, $i$ is the day where an intermediate image is needed ($i \in ]n;n+k[$). Once the image is computed, we add gaussian noise to the image $I_{n+i}$ to create another image $I_{G, (n+i)}$ including small variations as indicated in \autoref{eq:gaussian}. %The gaussian parameters were set to the mean $\mu = 0.04$ and standard deviation $\sigma = 0.02$. 
This noise is added to integrate randomness to the linear interpolation and avoid the ConvLSTM model to overfit the linear interpolation. Note that, the  \autoref{eq:interpolation} and \autoref{eq:gaussian} are element wise operation. 

\begin{equation}
\label{eq:interpolation}
I_{n+i} = \frac{I_{n} \times (k-i) + I_{n+k} \times k}{k} ; \ \ \ i \in ]n;n+k[
\end{equation}
\begin{equation}
\label{eq:gaussian}
I_{G, (n+i)} = I_{n+i} + \eta(\mu, \sigma)
\end{equation}

\subsubsection{ConvLSTM image generation}
The next phase consists in training the ConvLSTM model on the interpolated and real images, to generate a final dataset of daily artificial images. The ConvLSTM model is composed of four consecutive blocks, each one contains a \textit{ConvLSTM} layer, a \textit{BatchNorm} layer and a \textit{leakyReLU} layer. The model takes as input a set of daily ordered images: $\{D_{n-3}, D_{n-2}, D_{n-1}\}$ and outputs the next image, $D_{n}$. We train the model using the Adam optimizer and the Root Mean Square Error (RMSE) loss function. Once the model is trained, it is used to generate all the missing images and create a dataset for subsequent use. 

%The \textcolor{blue}{Root Mean Square Error (RMSE)} values seems good for the three vegetation indices used (see \ref{imagecoll} for more information about the vegetation indices).

\begin{equation}
\begin{split}
i_{t} = \sigma(W_{Di} \ast D_t + W_{hi} \ast h_{t-1} +  W_{ci} \circ C_{t-1} + b_i) \\
f_{t} = \sigma(W_{Df} \ast D_t + W_{hf} \ast h_{t-1} +  W_{cf} \circ C_{t-1} + b_f)  \\
C_t = f_t \circ C_{t-1} + i_t \circ \tanh(W_{Dc} \ast D_t + W_hc \ast h_{t-1} + b_c) \\
o_t = \sigma(W_{Do} \ast D_t + W_{ho} \ast h_{t-1} +  W_{co} \circ C_{t} + b_o) \\
h_t = o_t \circ \tanh{C_t}\\
\end{split}
\label{eq:convLSTM}
\end{equation}
The~\autoref{eq:convLSTM} illustrates the mathematical aspect of the ConvLSTM model where,  $W_{Di}$, $W_{hi}$, $W_{Df}$ ,$W_{hf}$ ,$W_{Dc}$ ,$W_{hc}$ , $W_{Do}$, $W_{ho}$, $W_{ci}$, $W_{cf}$ and $W_{co}$ denote the weights. $b_f$, $b_i$, $b_c$ and $b_o$ are the biases. The operation $\circ$ denotes element-wise product, $D_t$ is the current inputs and $h_{t-1}$ denotes the output of LSTM unit at the previous moment, and $\sigma$ as well as $\tanh(.)$ are nonlinear activation functions.

The ConvLSTM provides  intermediate images to fill the gap of the missing information. Using deep learning to create these images will enhance a basic linear interpolation, since the ConvLSTM can extract both temporal and spatial features within images.
%\textcolor{red}{
%\autoref{tab:convLstm_rmse} shows the results of the generated images compared to the ground truth images. Table
%\ref{imagecoll}}

\subsection{2D-1D fusion model}
\label{sec:fusion}
To combine satellite imagery with weather data, we propose  a fusion model based on transformers-encoders. Its architecture can be split into three main components: the ViT, the weather attention encoder, and a bottleneck transformer-encoder. The proposed architecture uses transformers to generate embeddings from the data, and then these embeddings are combined through the fusion bottleneck encoder. The output layer indicates whether crops are infected or not with a specific disease (here the downy mildew). \autoref{fig:process} illustrates the proposed approach. %The weather attention Encoder transforms chronological weather data into embeddings, while the ViT extracts features from the images. The concatenation of those features/embeddings is then analyzed by the last part of the model, the Fusion Bottleneck Encoder.

\subsubsection{Vision Transformer (ViT)}
\label{ssec:vitpart}
The ViT introduced by \cite{ViT} is based on a transformer-encoder. It becomes a popular architecture for its simplicity, and efficiency in terms of computation and performance. The ViT splits images in small patches ordered as a sequence, and uses only the encoder part of the traditional encoder-decoder transformer architecture, which makes it light. The transformer learns by measuring the relationship between image patches. This relationship can be learned by providing attention in the network. It is also possible to extract attention maps, which are useful for understanding the output of the model and the parts of the image on which the focus is. This is a key concept for anomaly detection, as we could interpret certain areas of the attention maps as infected crops.

%--------------------------------------- ViT Encoder Figure --------------------------------------%
\begin{figure}[H]
	\centering
	\includegraphics[width=1\linewidth]{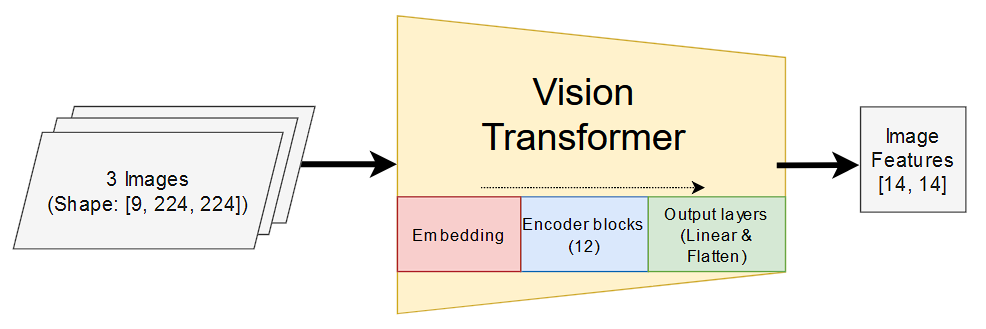}
	\caption{The ViT Encoder Architecture}
	\label{fig:ViT}
\end{figure}
%--------------------------------------- ViT Encoder Figure --------------------------------------%

The attention function is computed based on a set of queries simultaneously~\cite{Vaswani2017}, packed together into a matrix Q, where K and V denotes the keys and values of dimension $d_k$ \autoref{eq:attention}.
\begin{equation}
\label{eq:attention}
Attention(Q,K,V) = softmax\frac{QK^T}{\sqrt{d_k}}V
\end{equation}

\begin{equation}
\label{eq:multihead}
\begin{split}
MultiHeadAtt(Q,K,V) = \\Concat(Head_1, Head_2, ..., Head_h)W^o \\
Head_i = Attention(QW^Q_i,KW^K_i,VW^V_i)
\end{split}
\end{equation}
Multi-Head Attention is basically a linear projection of the queries, keys and values. %$h$ times with different learned linear projections to keys dimension $d_k$, queries dimension $d_k$ and values dimension $d_v$, respectively (\autoref{eq:multihead})\citet{Vaswani2017}.
VIT is based on three main components, patch embedding \autoref{eq:embedding}, feature extraction via stacked transformer encoders \autoref{eq:extraction} and classification head \autoref{eq:classification1} and \autoref{eq:classification2}.
\begin{equation}
\label{eq:embedding}
\begin{split}
z_0 = [x_{class}; x^1_pE; x^2_pE; ...; x^N_pE] + E_{pos} \\ 
\\
where \; E\in \R^{(P^2.C)\times D} , E_{pos}\in \R^{(N+1)\times D}
\end{split}
\end{equation}

\begin{equation}
\label{eq:extraction}
\begin{split}
z^{'}_l = MSA(LN(z_{l-1})) + z_{l-1}, \; where \;\; l = 1 ... L
\end{split}
\end{equation}

\begin{equation}
\label{eq:classification1}
z_l = MLP(LN(z'_l)) + z'_l, \; where \;\; l = 1 ... L
\end{equation}
\begin{equation}
\label{eq:classification2}
y = LN(z^0_L)   
\end{equation}
In the proposed architecture, the ViT takes inputs $I_V = \{x_1, x_2, x_3 \}$, where $x_i$ has a dimension of $(3, 224, 224)$ which represents a vegetation index  encoded with RGB image. Three different RGB vegetation indices, resulting in a tensor of size $(9, 224, 224)$. It outputs extracted features $O_V$ of dimension $(14, 14)$. \autoref{fig:ViT} illustrates the ViT encoder architecture used.
\\
\subsubsection{Weather Attention Encoder}

The Weather Attention Encoder is a simple Transformer Encoder that has the role of embedding the weather data. It takes as inputs $I_W = \{w_1,\dots, w_n\}$ a vector of weather features, such as rain duration and quantity, potential evaporation, sunlight amount, etc. First, the input $I_W$ is expanded to a high dimension using linear and dropout layers (\autoref{eq:1}). Then, this expanded vector is passed to the first encoder block (out of 12 encoder blocks) (\autoref{eq:2}). After these blocks, the output representation $O_W$ is produced using an network composed of linear, flatten and dropout layers (\autoref{eq:3}). We call this output $O_{W}$ the weather embeddings of dimension $(14, 14)$.

\begin{equation}
\label{eq:1}
A_W = Linear(Dropout(I_W))
\end{equation}
\begin{equation}
\label{eq:2}
A'_{W} = EncoderBlock(A_W)
\end{equation}
\begin{equation}
\label{eq:3}
O_W = Linear(Flatten(Dropout(A'_{W})))
\end{equation}
%\textcolor{red}{AH : ce n'est pas bien clair, encoder inside encoder,  vaut mieux ajouter un schéma}
%\textcolor{blue}{J'ai changé d'encoder à EncoderBlock. L'encoder est l'architecture globale, qui contient un quelques couches d'input, puis 12 blocks d'encoders (EncoderBlocks) puis un petit réseau de neurones en sortie. Je ne sais pas trop comment mieux expliquer, l'encoder contient 3 parties et une de ces parties est appellée "Encoder Block", et c'est quelque-chose que j'ai vu plusieurs fois sur des papiers de recherches et des repository Github donc je pense que c'est compréhensible.}

The encoder %blocks inside the encoder are classical encoder blocks, meaning is 
uses the architecture proposed by~\cite{Vaswani2017}. The encoder block can be decomposed in two sub-parts. The first part consists in multi-head attention and residual connections on the input vector $I_e$ to produce an intermediate features $H_e$ as indicated by \autoref{eq:enc:1}, on which layer normalization is applied (LN). The same process is made using a feed-forward network in the second sub-part of the encoder block described in \autoref{eq:enc:2}.

\begin{equation}
\label{eq:enc:1}
H_{e} = LN(I_{e} + Dropout(MultiHeadAtt(I_{e})))
\end{equation}
\begin{equation}
\label{eq:enc:2}
O_{e} = LN(H_{e} + Dropout(FFN(H_{e})))
\end{equation}
\\
\subsubsection{Fusion Bottleneck Encoder}

The Fusion Bottleneck Encoder takes as input $I_F$ the concatenation of the two previous embedded tensors that represents the visual and weather information, as described in \autoref{eq:4}. The concatenation ends up being a $(28, 14)$ tensor, since the $O_V$ and $O_W$ are of dimension $(14, 14)$ each. $I_F$ is then passed to the encoder, using the same process as the weather attention model. First, the vector is expanded (\autoref{eq:5}), then passed to the encoder (\autoref{eq:6}). The resulted $A'_F$ is then mapped to a vector using a feed-forward  network, made of linear, flatten and dropout layers ( \autoref{eq:7}). The last layer  $O_F$ consists of 2 outputs to predict whether there is disease of downy mildew or not (yes or no output).

\begin{equation}
\label{eq:4}
I_F = O_W \oplus O_V
\end{equation}
\begin{equation}
\label{eq:5}
A_F = Linear(Dropout(I_F))
\end{equation}
\begin{equation}
\label{eq:6}
A'_{F} = Encoder(A_F)
\end{equation}
\begin{equation}
\label{eq:7}
O_F = Linear(Flatten(Dropout(A'_{F})))
\end{equation}

%--------------------------------------- Fusion Bott. Figure --------------------------------------%
\begin{figure}[H]
	\centering
	\includegraphics[width=1\linewidth]{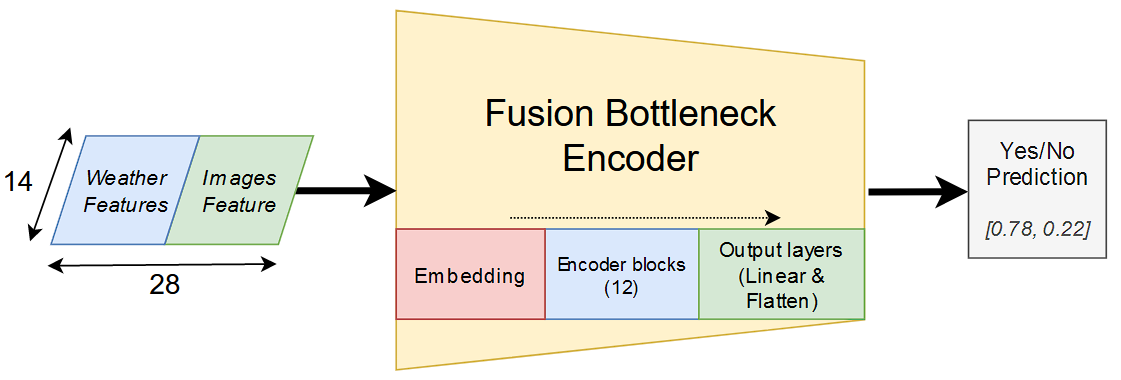}
	\caption{The Fusion Bottleneck Architecture}
	\label{fig:FBA}
\end{figure}
%--------------------------------------- Fusion Bott. Figure --------------------------------------%

%\subsubsection{Global Overview}

%From a global point of view, the model uses its sub-parts to generate embeddings for the data through encoders, and then uses these embeddings as inputs for the Fusion Bottleneck Encoder, deciding whether crops are infected or not. The Weather Attention Encoder transforms chronological weather data into embeddings, while the ViT extracts features from the images. The concatenation of those features/embeddings is then analyzed by the last part of the model, the Fusion Bottleneck Encoder. We give a figure representing the whole methodology process in the appendix (Figure \ref{fig:process}, Appendix \ref{appendix_1}).

%%%%%%%%%%%%%%%%%%%%%%%%%%%%%%%%%%%%%%%%%%%%%%%%%%%%%%%%%%%%%%%%%%%%%%%%%%%%%%%%%%%%%%%%%%

\section{Experiments and Results}
\label{experiments}
This section presents the different experiments conducted in this research work. It includes data collection, implementation details, evaluation results of the image generation and the fusion method. 

\subsection{Data collection}
Two types of data were collected, satellite images and weather conditions. 
\subsubsection{Image Collection}
\label{sec:imagecoll}
Remote sensing is an important data source for crop monitoring. The images of the vine crops from the studied site were collected using the Sentinel-2 Level-2A product. Sentinel-2 is a satellite mission of the European earth surveillance program  Copernicus. It provides multispectral images with different spatial resolutions 10, 20, and 60 m~\cite{cazaubiel2017multispectral}. The orthoimages were provided with atmospheric correction which is known to improve the images for subsequent use~\cite{Hadjimitsis2008}. In addition, Level-2A images have useful features such as cloud detection. 

%------------------------------------ Vegetation Indices Figure -----------------------------------%
\begin{figure*}[ht] 
	\centering
	\subfloat{%
		\includegraphics[width=0.2\linewidth]{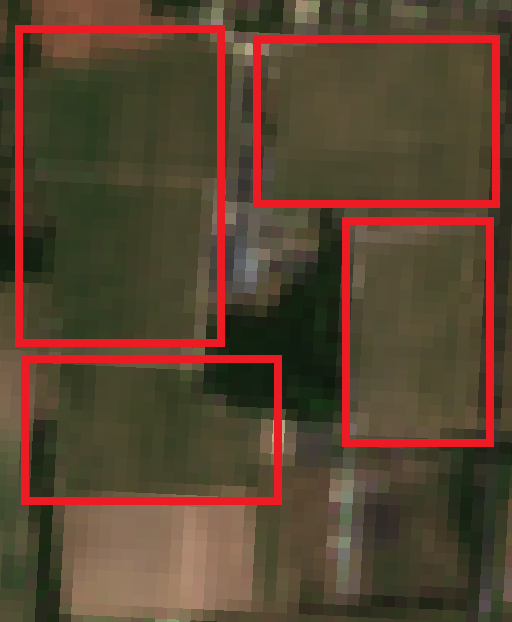} \label{1a}}
	\hfill
	\subfloat{%
		\includegraphics[width=0.2\linewidth]{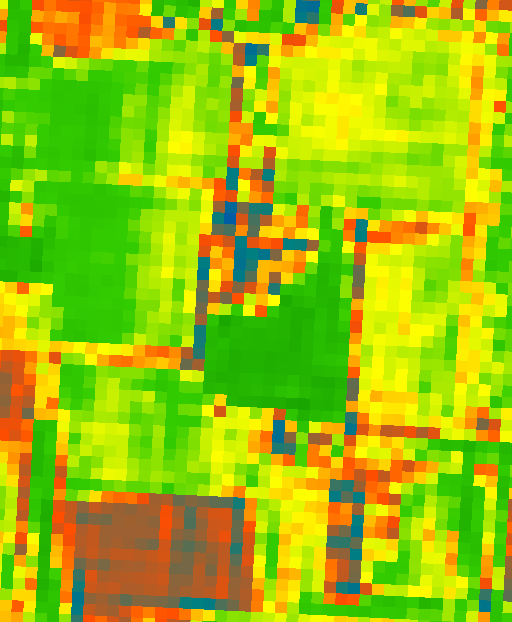}\label{1b}}
	\hfill
	\subfloat{%
		\includegraphics[width=0.2\linewidth]{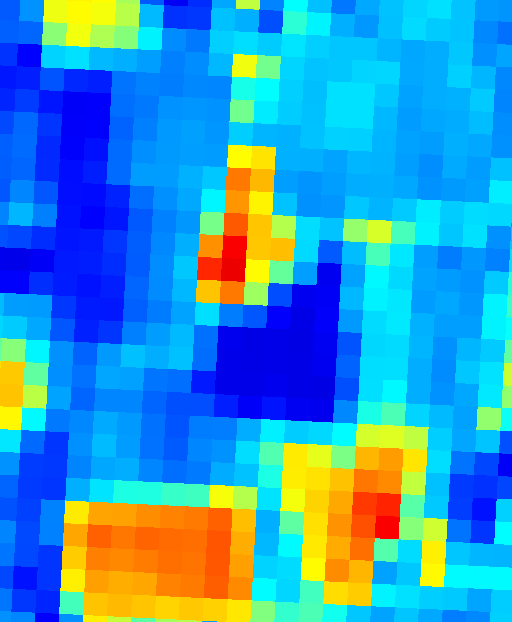}\label{1c}}
	\hfill
	\subfloat{%
		\includegraphics[width=0.2\linewidth]{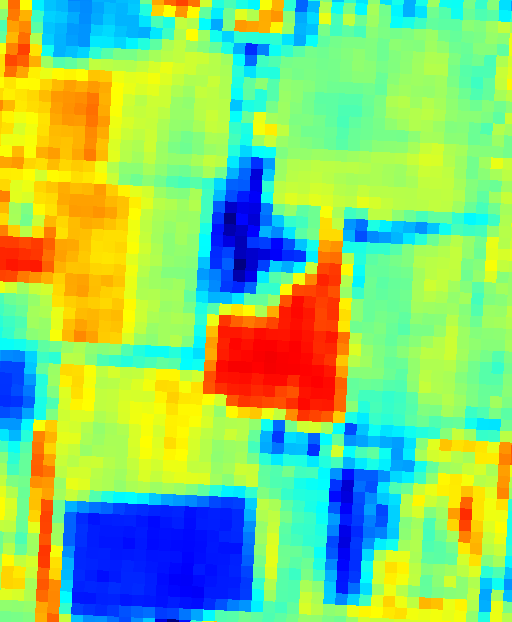}\label{1d}}
	\hfill
	\caption{Sample images for each vegetation index: (a) True Colors (studied crops outlined in red), (b) NDCI, (c) NDMI, (d) NDVI.}
	\label{fig:Indices} 
\end{figure*}

%------------------------------------ Vegetation Indices Figure -----------------------------------%
Sentinel-2 images provide rich sources of data related to the vegetation. The spectral bands allow the calculation of vegetation indices that are useful for measuring crop condition such as, vigor, biomass, chlorophyll content, and disease. 
The Difference Vegetation Index (NDVI) is a widely used vegetation index to measure the health status of vegetation, based on visible and near-infrared light reflected by vegetation. However, other type of indices can be useful for extracting more relevant information on vegetation or land condition. The Normalized Difference Chlorophyll Index (NDCI) introduced by~\cite{Mishra2012} is an index that describes the chlorophyll concentration, designed for water regions. The Normalized Difference Moisture Index (NDMI) measures the moisture levels of the crops and helps monitoring droughts. 
%\ place this : autoref{fig:Indices} shows images of the NDVI, and NDCI and NDMI indices of the studied vineyards zone.    \autoref{tab:2} lists the  Sentinel-2 bands of the three indices with their spatial resolution. 

In this study we used the three indices individually and in combination, all extracted from the SentinelHub API coded on RGB images. \autoref{tab:2} lists the  Sentinel-2 bands of the three indices with their spatial resolution.  \autoref{fig:Indices} shows the indices NDVI, and NDCI and NDMI extracted from the studied vineyards zone. 

%------------------------------------ Indices table Figure -----------------------------------%
\begin{table}[ht]{}
	\caption{Indices used in the study and their characteristics}
	\centering
	\begin{tabular}{p{1.5cm}p{2cm}p{2.5cm}}
		\hline
		\textbf{Index} & \textbf{Bands} & \textbf{Spatial Resolution}\\
		\hline 
		NDVI & B04, B08 & 10m\\
		NDCI & B04, B05 & 10m\\
		NDMI & B8A, B11 & 20m\\
		\hline
	\end{tabular}
	\label{tab:2}
\end{table}

In France, the downy mildew of vine generally starts to appear at the end of spring and begins to disappear with the fall of the leavesn~\cite{gessler_plasmopara_2011, vinopole_bordeaux-aquitaine_mildiou_nodate}.  To evaluate the presence of  downy mildew within the change in weather conditions for the same periods in each year, we selected images and weather data of June, July and August for the period 2018 until 2021 from the plots of {\em Lycée \,Agricole \,d'Amboise} in Centre Val de Loire region, France.

\subsubsection{Images Processing}
Even if the images are requested with a maximum cloud percentage of 10\%, there are still many images with large parts of clouds that slip through the filtering process. We deleted some of these remaining images using a cloud detection algorithm with pixel values and we cleaned the rest by hand. The satellite images cover a zone of four different vine plots. Since this zone contains buildings, trees, ... we cropped the images to only keep the appropriate vine parcels (see \autoref{fig:Indices} (a)).
In total, The dataset is composed of 1472 images (including the generated ones). We have 92 images per year per plot from June to August in four plots over four years from 2018 to 2021, where the number of real images per year are 38, 33, 40 and 29 respectively, more details are presented in \autoref{tab:datasetpartitionning}. Each image is labelled positive or negative to mildew depending on the ground truth. For our dataset, July and August of 2018 and 2021 were labelled as positive, the rest of the dataset were labeled negative).

\begin{table}[ht]
	\centering
	\caption{Dataset partitions}
	\begin{tabular}{@{}llllll@{}}
		\toprule
		\textbf{Year}           & 2018 & 2019 & 2020 & 2021 & \textbf{total} \\ \midrule
		\textbf{Real images}    & 38   & 33   & 40   & 29  &   \\
		\textbf{Total per plot} & 92   & 92   & 92   & 92  &  \\ 
		\textbf{Total images in all plots} & 368   & 386   & 368   & 368   & \textbf{1472}   \\ \bottomrule
	\end{tabular}
	\label{tab:datasetpartitionning}
\end{table}
% 2018 : 38 images réelles
% 2019 : 33 images réelles
% 2020 : 40 imùages réelles
% 2021 : 29 images réelles

\subsubsection{Weather Data}
The weather data was collected from the local meteorological station implemented on the studied site. It tracks daily weather conditions. The data contains: precipitation (in mm),  rainfall variability,  rainfall duration,  minimum, average and maximum temperature and humidity, potential evaporation, and on-site sun exposure.

\subsection{Implementation details and setup}
The experiments were performed on a computer with an Intel® Xeon(R) W-2123 CPU and an Nvidia GeForce GTX 1080 Ti, on Ubuntu 20.04 LTS. The model has been trained using the Adam optimizer with a learning rate of $1e-6$. We used a cosine warmup learning rate scheduler from the "transformers" library (\cite{Cosine_Warmup}), with a warmup of 100 epochs of 600 total epochs. The model early stopped  at the $120/130^{th}$ epoch.

The used hyperparameters for the ViT are: $12$ encoder-blocks layers with $8$ heads each, a patch size of $16$ pixels, an embedding size of dimension $768$, and images of size $224\times224$. For the other encoders of the model, we also use $12$ encoder-blocks layers using $8$ heads each and a feed-forward network size of $128$. The model embedding dimension is $64$.

%\subsection{Results}
Next section presents the evaluation results of the temporal image generation described in the section \ref{sec:Imgen}, and then the performance of the 2D-1D fusion architecture (section \ref{sec:fusion}) for downy mildew detection. We carried out the experimental study with the variation of several parameters of the proposed methods.

\subsection{Evaluation of the image generation method}
\begin{figure}[h]
	\centering
	\includegraphics[width=1\linewidth]{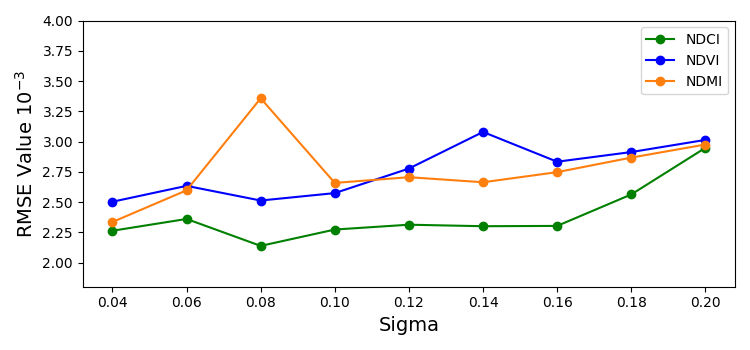}
	\caption{RMSE error according to different standard deviation of the noise added to the interpolated images of NDCI, NDVI and NDMI indices }
	\label{fig:RMSE_Sigma}
\end{figure}

The performance was evaluated with the one-leave-out cross-validation procedure. Each fold contains images of one year, so we have four folds for these experiments. The ConvLSTM network was trained on three folds and tested on one fold. Note that only the real images are involved in the test. Three previous images were used to predict the actual image. The RMSE metric measures the error between the predicted image and the real one. \autoref{tab:Imgen} shows the results obtained for each year, where the values are an average score on the set of images predicted by ConvLSTM trained with different noise values added to the interpolated images (\autoref{eq:gaussian}).  The noise tested is of Gaussian type with mean $\mu=0$ and different standard deviation $\sigma$, varying from 0.04 to 0.2, with a step of 0.02. We can observe in \autoref{tab:Imgen} that the RMSE is of the range of $10^{-3}$ for all indices, which represents a low prediction error.  The NDCI error is the smallest, followed by the NDVI and  NDMI. \autoref{fig:RMSE_Sigma} shows the RMSE error for the three indices as a function of the noise added to the interpolated images.  Similarly, the RMSE error remains low (range of $10^{-3}$) for the different values of the noise variance. We can notice some spikes in the curves, NDVI and NDMI. The RMSE of NDCI is more stable and smaller than the other two indices. Nevertheless, small noise values seem to work better for all indices. 
\begin{table}[ht]
	\centering
	\caption{Average of RMSE value for each indice and for each years over sigma values}
	\begin{tabular}{llll}
		\hline
		& NDCI                 & NDVI                 & NDMI     \\
		\hline
		
		\textbf{RMSE 2018 ($10^{-3}$)} & 1.94 & 2.23 & 2.15 \\
		\textbf{RMSE 2019 ($10^{-3}$)} & 2.95          & 3.071          & 3.86        \\
		\textbf{RMSE 2020 ($10^{-3}$)} & 2.18          & 2.67           & 2.42        \\
		\textbf{RMSE 2021 ($10^{-3}$)} & 2.47          & 3.07           & 2.64        \\   
		\textbf{Total ($10^{-3}$)}     & \textbf{2.38} & 2.75           & 2.76        \\
		\hline    
	\end{tabular}
	\label{tab:Imgen}
\end{table}

%\autoref{tab:convLstm_rmse} shows the results obtained by measuring the RMSE distance between the generated images and the ground truth.
\begin{figure}[h!]
	\centering
	\includegraphics[width=0.8\linewidth]{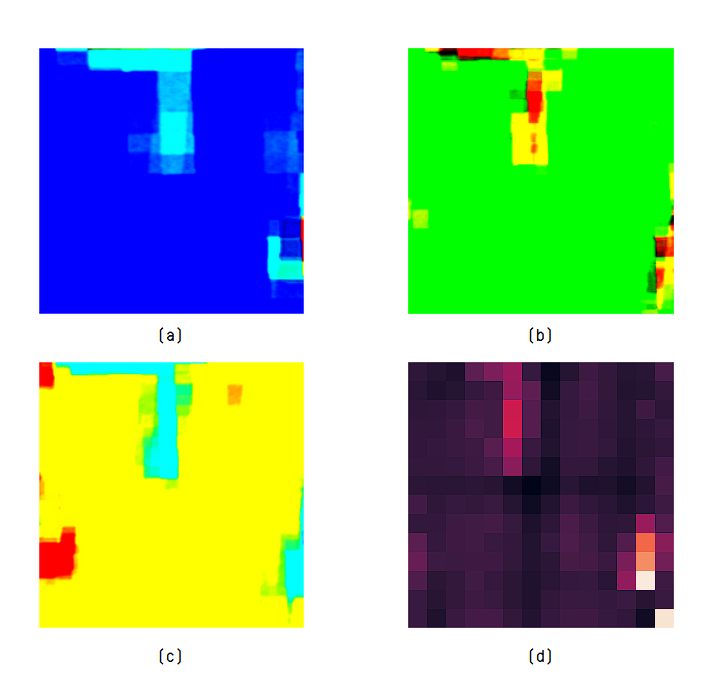}
	\caption{Three processed (resize, normalized, etc) vegetation indices and ViT attention map: (a) NDMI, (b) NDCI, (c) NDVI, (d) Attention map.}
	\label{fig:results}
\end{figure}
%------------------------------------ Tests Images Figure -----------------------------------%

\subsection{Evaluation of the 2D-1D fusion architecture}
The aim is to evaluate the performance of the proposed fusion method with different setups. We investigated the influence of combining different vegetation indices, ablation of some components, pruning of connections and changes in model layers. For this purpose, we used real images and those generated by the proposed method, as well as weather data, for training and testing. Data from 2018 (presence of downy mildew) and 2019 (no downy mildew) were considered for training. For the test, we used data from 2020 (no downy mildew), 2021 (downy mildew). The performance measurement was performed by the well known accuracy measure and the F1 score.

%------------------------------------ Indices Combination Figure -----------------------------------%
\begin{table*}[h]
	\caption{Evaluation of the fusion method with different combinations of vegetation indices}
	\centering
	\begin{tabular}{p{2.5cm}p{2.5cm}p{2cm}p{2cm}p{2cm}p{1cm}p{1cm}p{1cm}}
		\hline
		\textbf{    } & \textbf{NDCI, NDVI, NDMI} & \textbf{NDVI, NDMI} & \textbf{NDCI, NDMI} & \textbf{NDCI, NDVI} & \textbf{NDVI} & \textbf{NDCI} & \textbf{NDMI}\\
		\hline 
		\textbf{Accuracy} & \textbf{0.970}	& 0.912	 & 0.641 & 0.913 & 0.706 & 0.956 & 0.751\\
		\textbf{F1-Score} & \textbf{0.975} &	0.924 &	0.771 &	0.928 &	0.802 &	0.966 &	0.816\\
		\textbf{Test loss} & \textbf{0.03} &	0.176 &	0.657 &	0.052 &	0.565 &	0.561 &	0.278\\
		\hline 
	\end{tabular}
	\label{tab:3}
\end{table*}
%------------------------------------ Indices Combination Figure -----------------------------------%

\subsubsection{Vegetation Indices Combinations}
Multiple vegetation indices combination were tested, the best accuracy and F1-Score were obtained with the combination of all three indices. As we can observe in the \autoref{tab:3}, NDCI might be the one playing the most important role as it produces a very good accuracy on its own and performs well with both NDVI and NDMI. Even though this vegetation index seem to be the most important, we cannot neglect others as the combination of all three outperforms by more than 2\%.

\autoref{fig:results} shows vegetation indices images used as an input for the ViT part of the model (section \ref{ssec:vitpart}). We can observe that the attention map looks similar to the vegetation indices images, as it highlights the shape of some parts of the image that may be related to downy mildew. This also indicates that the model seems to learn to recognize patterns in vegetations indices and that it's not highlighting uninteresting parts of the image.

\subsubsection{Ablation Study}
To evaluate whether the network architecture 
is biased by the weather features or the image features, we tested the network with an ablation method to cut off parts of the network. This consists in cutting one of the parts of the network, either the ViT encoder or the Weather Attention Encoder. We replace the concatenation of weather and image features of dimension $(28, 14)$ by a $(14, 14)$ features map as if the other part of the model did not exist, and we adapt the Fusion Bottleneck Transformer to use these new dimensions as input by tweaking the input parameters. We then train this new model using the same process as for the global model, ignoring the missing parts. 

%replaced Regular model -> our model

%------------------------------------ Ablation Study Figure -----------------------------------%
\begin{table}[ht]
	\caption{Results of the ablation study of the fusion model}
	\centering
	\begin{tabular}{p{1.5cm}p{1.7cm}p{1.7cm}p{1.7cm}}
		\hline
		\textbf{    } & \textbf{Image only} & \textbf{Weather Only} & \textbf{Fusion Model}\\
		\hline 
		Accuracy & 0.481 & 0.662 & \textbf{0.970}\\
		F1-Score & 0.630 & 0.790 & \textbf{0.975}\\
		\hline
	\end{tabular}
	\label{tab:4}
\end{table}
%------------------------------------ Ablation Study Figure -----------------------------------%

After the training, we obtain the following results, described in the \autoref{tab:4}. When cutting off the weather data, the model achieves 48.1\% accuracy, which is fairly bad, and when cutting the images part, the accuracy goes up to 66.2\%; which is better but still not good. What this study shows is that the model does not seem to be biased by either weather or images, and that it needs both to learn correctly.

\subsubsection{Pruning Study}

Network pruning has become common to study models, as it can enlighten a potential overparameterization. We tested pruning by taking the parameters that give the best accuracy (97\% accuracy and 97.5\% F1-Score), and we prune parameters of either part of the network (only Weather Attention Encoder for instance) or the whole network.

%------------------------------------ Pruning Study Figure -----------------------------------%
\begin{table}[h]
	\caption{Results of the fusion model pruning. W.A.E. denotes Weather Attention Encoder, ViT denotes Vision Transformers and Fusion B. denotes Fusion Bottelneck and Global means pruning global}
	\centering
	\begin{tabular}{p{1.3cm}p{1.3cm}p{1.3cm}p{1.3cm}p{1.3cm}}
		\hline
		\textbf{\% pruned} & \textbf{W.A.E.} & \textbf{Vision T.} & \textbf{Fusion B.} & \textbf{Global}\\
		\hline 
		1\%	& 0.733 &	\textbf{0.982} &	0.555 &	0.516\\
		5\%	& 0.647 &	0.976 &	0.54 &	0.478\\
		7.5\%& 	0.635 &	0.969 &	0.461 &	0.508\\
		10\%&	0.612&	0.962&	0.543&	0.501\\
		\hline
	\end{tabular}
	\label{tab:5}
\end{table}
%------------------------------------ Pruning Study Figure -----------------------------------%

\autoref{tab:5} presents the accuracy of the model when pruning 1\%, 5\%, 7.5\% and 10\% of each model part. We can observe that the ViT is more resilient to these prunings, and it can be explained by its number of parameters; the ViT has approximately 86 millions parameters which makes it robust, while the other encoders of the model have almost three times less parameters. As we saw in the ablation study, keeping only the weather gives better results than keeping only the images, and here we can see that removing parts of the ViT does not affect the accuracy by a lot. This means our model gives the best results when using images with weather, but the weather module is still the most important part of the network.

\subsubsection{Encoders layers}
The number of encoder layers impacts the accuracy of the model as shown in the Figure \ref{fig:layers} below. We consider 12 layers the best number for the model as it gives the best results while maintaining a fast training.

%------------------------------------ Layers Figure -----------------------------------%
\begin{figure}[h!]
	\centering
	\includegraphics[width=0.9\linewidth]{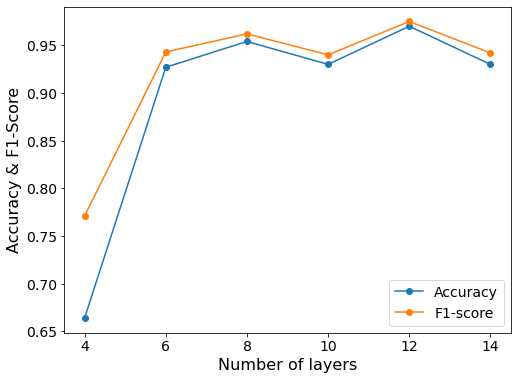}
	\caption{Model precision depending on number of encoder layers}
	\label{fig:layers}
\end{figure}
%------------------------------------ Layers Figure -----------------------------------%

%%%%%%%%%%%%%%%%%%%%%%%%%%%%%%%%%%%%%%%%%%%%%%%%%%%%%%%%%%%%%%%%%%%%%%%%%%%%%%%%%%%%%%%%%%

\section{Discussions}
The first hypothesis of this research states that a deep neural network trained with data obtained by linear interpolation will be able to generate artificial images intermediate between two temporally distant acquisitions. Thus allowing to match and merge heterogeneous data acquired with different frequencies. In this research, it was found that interpolation combined with a deep network (ConvLSTM) provides good results to fill the information gap in satellite images. The error is small when the generated images are compared to the real ones. Another important result is that perturbing the linear interpolation with Gaussian noise, helps to improve the training and convergence of the ConvLSTM network. Therefore, we can assume that this method could be an interesting tool to complete the data and allow better performance of deep networks. 

Although ConvLSTM achieves good RMSE. 
Further research should be undertaken to study the effectiveness of this model with more data. Real satellite images (those that are not artificially generated) are too infrequent and there is no guarantee that the intermediate images are the ground truth. Nevertheless, given the results obtained, we can expect that the images generated will be close to the real images, and this will not significantly affect the machine learning process. A another point, the artificial generation of intermediate images could also induce noise in the images or pixels invisible to the naked eye, which could bias the network. This is certainly not the case because we can observe on \autoref{tab:4} that when the ViT is cut, the model still seems able to learn, even if it is not very accurate. 

The second working hypothesis consisted in showing the effectiveness of transformers for the implementation of information fusion with the aim of crop disease detection.  The difficulty lies in the heterogeneity of the data and their dimensions, as well as the difference in the acquisition frequencies. The proposed method was able to address these problems. The obtained scores showed that the fusion of multispectral satellite images (2D) and weather data (1D) offers better performances for the detection and identification of downy mildew. Indeed, for the test set, our model outputs predictions with a 97\% accuracy and a 97.5\% F1-Score. These results seem promising, and the qualitative analysis of attention over images seem to confirm that the network learnt how to recognize downy mildew on vegetation indices of satellites images.

The weather has an impact on the development of downy mildew, the vegetation indices include indications on the health status of a crop. The results showed that the combination of the two parts gives better scores than a single type of data. Comparing the results of ablation experiments shows that each module (weather data, image), affects the fusion results, as depicted in \autoref{tab:4}. From the results obtained we can assume that each module provides complementary information which increases significantly the performances. It was also found that using all three vegetation indices (NDVI, NDCI, NDMI) is more effective than one or two. Nevertheless, it seems that NDCI has a little more discriminating elements compared to the other two (see \autoref{tab:3}).  

The proposed fusion model performs well, but there is room for improvement on several aspects of the methodology. For instance, improvement could be made on the fusion model. It seems that the model gains a little bit in accuracy when a small percentage of its parameters are cut, which possibly indicates redundancies in the parameters. We believe it is possible to modify the Weather Attention Encoder and Fusion Bottleneck Encoder to enhance them for this task, using a custom loss function or small modifications in the architecture.

%%%%%%%%%%%%%%%%%%%%%%%%%%%%%%%%%%%%%%%%%%%%%%%%%%%%%%%%%%%%%%%%%%%%%%%%%%%%%%%%%%%%%%%%%%

\section{Conclusion}
In this research work, the objective was to take advantage of the complementarity of heterogeneous data to detect downy mildew disease in vine plots. For this purpose, we have proposed a new framework where the first part consists in generating intermediate satellite images,  to create daily images and pairing them with the weather data. The second part consists of a new data fusion architecture based on transformer networks that combine weather data and satellite images. We used two transformers (an encoder and ViT) to extract modalities features/embeddings (weather and satellite images), and a third bottleneck transformer-encoder outputs the prediction based on the embeddings/features. We achieve promising results which validates the hypothesis that fusion using multiple transformers instead of a single custom one is possible and can achieve good results. We believe these results are interesting for agriculture challenges as it could lead to better crops monitoring. Also the proposed architecture can be used for other application domains. 

%%%%%%%%%%%%%%%%%%%%%%%%%%%%%%%%%%%%%%%%%%%%%%%%%%%%%%%%%%%%%%%%%%%%%%%%%%%%%%%%%%%%%%%%%%

\section{Acknowledgment}
This work was carried out as a part of MERIAVINO European project. We gratefully acknowledge ERA-NET ICT-AGRI-FOOD, and national research agencies: ANR, UEFISCDI and GSRI for their support. We thank Institut Francais de la Vigne (IFV) for providing ground truth information, and Lycée Agricole d'Amboise for providing studied plots. 

%%%%%%%%%%%%%%%%%%%%%%%%%%%%%%%%%%%%%%%%%%%%%%%%%%%%%%%%%%%%%%%%%%%%%%%%%%%%%%%%%%%%%%%%%%

 \bibliographystyle{IEEEtran}
 \bibliography{ref}

\end{document}